\documentclass{article}

\usepackage{microtype}
\usepackage{graphicx}
\usepackage{subcaption}
\usepackage{booktabs} 

\usepackage{hyperref}

\usepackage[accepted]{icml2024}

\usepackage{amsmath}
\usepackage{amssymb}
\usepackage{mathtools}
\usepackage{amsthm}

\usepackage[capitalize,noabbrev]{cleveref}

\theoremstyle{plain}
\newtheorem{theorem}{Theorem}[section]

\newtheorem{lemma}[theorem]{Lemma}
\newtheorem{corollary}[theorem]{Corollary}
\theoremstyle{definition}
\newtheorem{definition}[theorem]{Definition}

\theoremstyle{remark}
\newtheorem{remark}[theorem]{Remark}

\usepackage[textsize=tiny]{todonotes}

\newcommand\bfr{\mathbf{r}}
\newcommand\Reg{\mathsf{Reg}}
\newcommand\cA{\mathcal{A}}
\newcommand\cP{\mathcal{P}}
\newcommand\cN{\mathcal{N}}
 
\DeclareMathOperator*{\expect}{\mathbb{E}}
\newcommand{\E}[2]{ 
\ifthenelse{\isempty{#1}}{\expect\left[{#2}\right]}{\expect_{#1}\left[{#2}\right]}
}
\newcommand{\eps}{\epsilon} 

\icmltitlerunning{Online Learning with Bounded Recall}

\begin{document}

\twocolumn[
\icmltitle{Online Learning with Bounded Recall}



\icmlsetsymbol{equal}{*}

\begin{icmlauthorlist}
\icmlauthor{Jon Schneider}{equal,comp}
\icmlauthor{Kiran Vodrahalli}{equal,comp}
\end{icmlauthorlist}

\icmlaffiliation{comp}{Google}

\icmlcorrespondingauthor{Jon Schneider}{jschnei@google.com}
\icmlcorrespondingauthor{Kiran Vodrahalli}{kirannv@google.com}

\icmlkeywords{online learning, bounded recall, no-regret}

\vskip 0.3in
]



\printAffiliationsAndNotice{\icmlEqualContribution} 

\begin{abstract}
We study the problem of full-information online learning in the ``bounded recall'' setting popular in the study of repeated games.  An online learning algorithm $\mathcal{A}$ is $M$-\textit{bounded-recall} if its output at time $t$ can be written as a function of the $M$ previous rewards (and not e.g. any other internal state of $\cA$). We first demonstrate that a natural approach to constructing bounded-recall algorithms from mean-based no-regret learning algorithms (e.g., running Hedge over the last $M$ rounds) fails, and that any such algorithm incurs constant regret per round. We then construct a stationary bounded-recall algorithm that achieves a per-round regret of $\Theta(1/\sqrt{M})$, which we complement with a tight lower bound. Finally, we show that unlike the perfect recall setting, any low regret bound bounded-recall algorithm must be aware of the ordering of the past $M$ losses -- any bounded-recall algorithm which plays a symmetric function of the past $M$ losses must incur constant regret per round.
\end{abstract}

\section{Introduction}

Online learning is the study of online decision making, and is one of the cornerstones of modern machine learning, with numerous applications across computer science, machine learning, and the sciences. 

Traditionally, most online learning algorithms depend (either directly or indirectly) on the entire history of rewards seen up to the present. For example, each round, the Multiplicative Weights Update method decides to play action $i$ with probability proportional to some exponential of the cumulative reward of action $i$ until the present (in fact, the rewards of the first round have as much impact on the action chosen at round $t$ as the rewards of round $t-1$). 

Another natural setting for learning is to restrict the learner so that they can only use information they received from the past $M$ rounds of play. This is a restriction that is known in the game theoretic literature as \emph{bounded recall}, and was introduced by \citet{aumann1989cooperation} in an attempt to design a model of repeated play that could lead to altruistic equilibria. However, there are multiple other reasons why bounded recall is interesting to study from a learning angle, including:

\begin{itemize}

\item \textbf{Recency bias and modelling human behavior}: It is increasingly common in the economics and social sciences literature to model rational agents as low-regret learners (e.g. \citet{blum2007learning,nekipelov2015econometrics,braverman2018selling}). However, the extent to which standard low-regret algorithms actually model human behavior (especially over longer timescales) is unclear. Indeed, ``recency bias'' is a well-known psychological phenomenon in human decision-making indicating that people do, in general, prioritize more recent information. Bounded recall dynamics could be a more accurate model of human behavior than general no-regret dynamics.

\item \textbf{Designing adaptive learning algorithms:} On the flip side, one may wish to design learning algorithms that \emph{do} prioritize information from more recent rounds. While there exist learning paradigms that accomplish this without imposing the bounded-recall constraint (notably, algorithms that minimize adaptive / dynamic regret: see \citet{bousquet2002tracking, zhang2018dynamic, zheng2019equipping}), it is also valuable to understand the black box procedure of running an arbitrary learning algorithm over a sliding window.

\item \textbf{Sequence models:} Decisions made by attention-based sequence models \citep{vaswani2017attention}, which are popularly employed by modern large language models \citep{achiam2023gpt, team2023gemini}, fall naturally into the bounded recall paradigm: these models have a fixed context window of past rounds (corresponding to tokens) that they can observe, but the models can also be used to operate on streams of tokens far exceeding their context window. As decisions are increasingly entrusted to sequence models, it is important to understand their theoretical capabilities as online learning algorithms.

\item \textbf{Privacy and data retention}: Finally, the rights of users to control the use of their data are becoming ever more common worldwide, with regulations like the GDPR's ``right to be forgotten'' \citep{voigt2017eu} influencing how organizations store and think about data. For instance, it is increasingly recognized that it is not enough to simply remove explicit storage of data, but it is also necessary to remove the impact of this data on any models that were trained using it (motivating the study of machine unlearning, see e.g. \citet{ginart2019making}).

Similarly, many organizations have a blanket policy of erasing any data that lies outside some retention window; such organizations must then train their model with data that lies within this window. Note that this is an example of a scenario where it is not sufficient to simply prioritize newer data; we must actively exclude data that is sufficiently old.
\end{itemize}

Motivated by these examples, in this paper we study \emph{bounded-recall} learning algorithms: algorithms that can only use information received in the last $M$ rounds when making their decisions. In particular, we study this problem in the classical full-information online learning model, where every round $t$ (for $T$ rounds) a learner must play a distribution $x_t$ over $d$ possible actions\footnote{Throughout this paper we focus entirely on the learning with experts setting (where the action set is the $d$-simplex), but all observations / theorems in this paper should extend easily to other full-information online learning settings, such as online convex optimization. It is an interesting question to understand whether it is possible to adapt these results for partial-information settings such as bandits.}. Upon doing this, the learner receives a reward vector $r_t$ from the adversary, and receives an expected reward of $\langle x_t, r_t \rangle$. The learner's goal is to minimize their regret (the difference between their utility and the utility of the best fixed action).

It is well-known that without any restriction on the learning algorithm, there exist low-regret learning algorithms (e.g. Hedge) which incur at most $O(\sqrt{(\log d)/T})$ regret per round, and moreover this is tight; any algorithm must incur at least $\Omega(\sqrt{(\log d)/T})$ regret on some online learning instance. We begin by asking what regret bounds are possible for bounded-recall algorithms. In Section \ref{sec:benchmarks} we show that (in analogy with the unrestricted case), any bounded-recall learning algorithm must incur at least $\Omega(\sqrt{(\log d)/M})$ regret per round. Moreover, there is a very simple bounded-recall algorithm which achieves this regret bound: simply take an optimal unrestricted learning algorithm and restart it every $M$ rounds (Algorithm \ref{alg:restart_NR}). 

However, this periodically restarting algorithm has some undesirable qualities -- for example, even when playing it on time-invariant stochastic data, the performance of this algorithm periodically gets worse every $M$ rounds right after it restarts (it also e.g. does not seem like a particularly natural algorithm for modeling human behavior). Given this, we introduce the notion of \emph{stationary} bounded-recall algorithms, where the action taken by the learner at time $t$ must be a fixed function of the rewards in rounds $t-M$ through $t-1$; in particular, this function cannot depend on the value $t$ of the current round (and rules out the periodically restarting algorithm).

One of the most natural methods for constructing a stationary bounded-recall algorithm is to take an unrestricted low-regret algorithm $\cA$ of your choice and each round, rerun it over only the last $M$ rewards. Does this procedure result in a low-regret stationary bounded-recall algorithm? We prove a two-pronged result:

\begin{itemize}
    \item If the low-regret algorithm $\cA$ belongs to a class of algorithms known as \textit{mean-based algorithms}, then there exists an online learning instance where the resulting bounded-recall algorithm incurs a \textit{constant} (independent of $M$ and $T$) regret per round.
    
    The class of mean-based algorithms includes most common online learning algorithms (including Hedge and FTRL); intuitively, it contains any algorithm which will play action $i$ with high probability if historically, action $i$ performs linearly better than any other action.
    
    \item However, there \textit{does} exist a low-regret algorithm $\cA$ such that the resulting stationary bounded-recall algorithm incurs an optimal regret of at most $O(\sqrt{(\log d)/M})$ per round. 
    
    We call the resulting bounded-recall algorithm the ``average restart'' algorithm and it works as follows: first choose a random starting point $s$ uniformly at random between $t-M$ and $t-1$. Then, run a low-regret algorithm of your choice (e.g. Hedge) only on the rewards from rounds $s$ to $t-1$. 
    
    The underlying full-horizon algorithm can be obtained by setting $M = T$. Interestingly, as far as we are aware this appears to be a novel low-regret (non-mean-based) algorithm, and may potentially be of use in other settings where mean-based algorithms are shown to fail (e.g. \citet{braverman2018selling}).

    \item In contrast with standard, full-horizon multiplicative weights (which assign equal importance to all previous rounds), the ``average restart'' algorithm we construct treats the set of rounds it can observe highly asymmetrically, putting far more weight on more recent rounds. We show that this asymmetry is essential in the bounded recall setting: any symmetric, stationary bounded-recall algorithm must incur high per-round regret.
\end{itemize}

Finally, we run simulations of the algorithms mentioned above on a variety of different online learning instances. We observe that in time-varying settings, the bounded-recall algorithms we develop often outperform their standard online learning counterparts.

\paragraph{Related work.} 
The idea of bounded-recall dynamics in economics appears to have been introduced by either \citet{aumann1989cooperation} or \citet{lehrer1988repeated}. Since then, there has been a large economic literature on studying games under bounded-recall dynamics and repeated games where agents have bounded memory (e.g., with strategy encoded by a finite-state machine). See \citet{neyman1997cooperation} for a partial survey of the area. See also \citet{drenska2023online} for a related but different problem of getting low regret compared to a class of benchmarks which are bounded recall (instead of designing a learning algorithm which must be bounded recall). \citet{qiao2021exponential} have also considered a bounded-recall setup in the selective learning problem setting. Most relevant to this paper is the work of \citet{zapechelnyuk2008better}, who shows (in a similar proof to our Theorem \ref{thm:main_counterexample}) that ``better-reply dynamics'' in bounded recall settings can lead to high regret (``better-reply dynamics'' are a generalization of regret matching algorithms, and are a more constrained class than the set of mean-based algorithms we study). The fact that \textsc{PeriodicRestart} achieves low-regret is a folklore result in both online learning and repeated games. To the best of our knowledge, the algorithms \textsc{AverageRestart} and \textsc{AverageRestartFullHorizon} have not been considered before in either literature.

We defer discussion of additional related work to Appendix \ref{sec:related_work}.

\section{Model and Preliminaries}

We consider a full-information online learning setting where a learner (running algorithm $\mathcal{A}$) must output a distribution $x_t \in \Delta([d])$ over $d$ actions each round $t$ for $T$ rounds. Initially, an oblivious\footnote{Note that all results we obtain for our deterministic algorithms immediately extend to the adaptive setting, since in the full-information setting an oblivious adversary can already perfectly predict what we are going to play in any round.} adversary selects a sequence $\bfr \in [0,1]^{T\times d}$ of reward vectors $\{r_t\}_{t \in [T]}$ where $r_{t, i}$ represents the reward of action $i$ in round $t$. Then, in each round $t$, the learner selects their distribution $x_t \in \Delta([d])$ as a function of the reward vectors $r_1, r_2, \dots, r_{t-1}$ (we write this as $x_t = \cA(r_1, r_2, \dots, r_{t-1})$). The learner then receives utility $\langle x_t, r_t\rangle$, and observes the full reward vector $r_t$. We evaluate our performance via the \textit{per-round regret}:

\begin{definition}[Per-round Regret]
The per-round regret of an online learning algorithm $\mathcal{A}$ on a learning instance $\bfr$ is given by
\[
\Reg(\mathcal{A}; \bfr) := \frac{1}{T}\left(\max_{i \in [d]} \sum_{t = 1}^T r_{t, i} - \sum_{t = 1}^T \langle x_t, r_t\rangle\right).
\]
In other words, $\Reg(\mathcal{A}; \bfr)$ represents the (amortized) difference in performance between algorithm $\mathcal{A}$ and the best action in hindsight on instance $\bfr$. Where $\bfr$ is clear from context we will omit it and write this simply as $\Reg(\mathcal{A})$. 
\end{definition}

Throughout this paper we will primarily be concerned with online learning algorithms that are \textit{bounded-recall}; that is, algorithms whose decision can only depend on a subset of recent rewards. Formally, we define this as follows:

\begin{definition}[Bounded-Recall Online Learning Algorithms]\label{def:bounded-recall}
An online learning algorithm $\mathcal{A}$ is \emph{$M$-bounded-recall} if its output $x_t$ at round $t$ can be written in the form 

$$x_t = f_t(r_{t - M}, \dots, r_{t-1})$$

\noindent
for some fixed function $f_t$ depending only on $\mathcal{A}$ (here we take $r_i = 0$ when $i \leq 0$); in other words, the output at time $t$ depends only on $t$ and the rewards from the past $M$ rounds (the \emph{history window}). If furthermore we have that $f_t$ is independent of $t$, we say that $\mathcal{A}$ is a \emph{stationary $M$-bounded-recall} algorithm. 
\end{definition}

In general, we will consider the setting where both $M$ and $T$ go to infinity, with $T \gg M$ (although many of our results still hold for $T = \Theta(M)$). We say a ($M$-bounded-recall) learning algorithm is \textit{low-regret} if $\Reg(\mathcal{A}) = o(1)$ (here we allow $o(1)$ to depend on $M$; i.e., we will consider an algorithm with $\Reg(\mathcal{A}) = O(1/\sqrt{M})$ to be low-regret). 

\subsection{Mean-based learners}

One of the most natural approaches to constructing a stationary bounded-recall learner $\mathcal{A}$ is to take a low-regret learning algorithm $\mathcal{A'}$ for the full-horizon setting (e.g. the Hedge algorithm) but only run it over the most recent $M$ rounds. Later, we show that for a wide range of natural low-regret learning algorithms $\mathcal{A'}$ -- e.g., Hedge, follow the perturbed/regularized leader, multiplicative weights, etc. -- this results in a bounded-recall algorithm $\mathcal{A}$ with \textit{linear regret}.

All these algorithms have the property that they are \textit{mean-based} algorithms. Intuitively, an algorithm is mean-based if it approximately best responds to the history so far (i.e., if one arm $i$ has historically performed better than all other arms, the learning algorithm should play $i$ with weight near $1$). Formally, we define this as follows.

\begin{definition}[Mean-based algorithm]
For each $i \in [d]$ and $1 \leq t \leq T$, let $R_{t, i} = \sum_{s=1}^{t} r_{s, i}$. A learning algorithm $\mathcal{A}$ is \emph{$\gamma$-mean-based} if, whenever $R_{t, i} - R_{t, j} > \gamma T$, $x_{t, j} < \gamma$. A learning algorithm is \textit{mean-based} if it is $\gamma$-mean-based for some $\gamma = o(1)$. 
\end{definition}

\citet{braverman2018selling} show that many standard low-regret algorithms (including Hedge, Multiplicative Weights, Follow the Perturbed Leader, EXP3, etc.) are mean-based.

We say a $M$-bounded-recall algorithm is \textit{$M$-mean-based} if its output $x_t$ in round $t$ is of the form $\mathcal{A}_t(r_{t-M}, \dots, r_{t-2}, r_{t-1})$ for some mean-based algorithm $\mathcal{A}_t$ (note that we allow for the choice of mean-based algorithm to differ from round to round; if we wish to construct a stationary $M$-bounded-recall algorithm, $\mathcal{A}_t$ should be the same for all $t$). 

\section{Benchmarks for bounded-recall learning}
\label{sec:benchmarks}

We begin by showing that all $M$-bounded-recall learning algorithms must incur at least $\Omega(\sqrt{(\log d)/M})$ regret per round. Intuitively, this follows for the same reason as the $\Omega(1/\sqrt{T})$ regret lower bounds for ordinary learning: a learner cannot distinguish between reward signals with mean $1/2$ and with mean $1/2 + \sqrt{1/M}$ with $o(M)$ samples. 

\begin{theorem}[Lower Bound]
\label{thm:lb}
Fix an $M > 0$. Then for any $M$-bounded-recall learning algorithm $\mathcal{A}$ and $T > M$, there exists a distribution $\mathcal{D}$ over online learning instances $\bfr{}$ of length $T$ with $d$ actions such that

$$\mathbb{E}_{\bfr \sim \mathcal{D}}[\Reg(\mathcal{A}; \bfr{})] \geq \Omega\left(\sqrt{\frac{\log d}{M}}\right).$$
\end{theorem}
\begin{proof}
See Appendix.
\end{proof}

Next, we show that we can achieve the regret bound of Theorem \ref{thm:lb} with a very simple bounded-recall algorithm family which we call the \textsc{PeriodicRestart} algorithm (Algorithm \ref{alg:restart_NR}). 

\begin{algorithm}
\caption{\textsc{PeriodicRestart}}\label{alg:restart_NR}
\begin{algorithmic}[1]
\REQUIRE Time horizon $T$, history window $M$, no-regret algorithm $\mathcal{A}$.
\FOR{$t = 1 \to T$} 
\STATE{$s := \lfloor t/M \rfloor \cdot M$}
\STATE{Play action $x_t = \mathcal{A}(r_{s}, r_{s+1}, \dots, r_{t})$.}
\ENDFOR
\end{algorithmic}
\end{algorithm}

Note that in \textsc{PeriodicRestart}, $x_t$ depends on at most the previous $M$ rounds and $t$. Note too that it is straightforward to implement \textsc{PeriodicRestart} given an implementation of $\cA$; it suffices to simply run $\cA$, restarting its state to the initial state every $M$ rounds (hence the name of the algorithm). 

\begin{theorem}\label{thm:periodic_regret}
Assume the algorithm $\mathcal{A}$ has the property that $\Reg(\mathcal{A}; \bfr) \leq R(T, d)$ for any online learning instance $\bfr \in [0, 1]^{T \times d}$. Then, for any instance $\bfr \in [0, 1]^{T \times d}$

$$\Reg(\textsc{PeriodicRestart}; \bfr) \leq R(M, d).$$
\end{theorem}
\begin{proof}
We will show that the guarantees on $\cA$ imply that the per-round regret of \textsc{PeriodicRestart} over a segment of length $M$ where $\cA$ does not restart is at most $R(T, d)$ (from which the theorem follows). Let $i^* = \arg\max_{i\in [d]} \sum_{t} r_{t, i}$. Since $\Reg(\cA; \bfr) \leq R(T, d)$, for any $0 \leq n \leq \lfloor T/M \rfloor$, it is the case that

\begin{eqnarray*}
\sum_{t=nM + 1}^{(n+1)M} \langle x_t, r_t \rangle &=& \sum_{t=nM+1}^{(n+1)M} \langle \cA(x_{nM+1}, x_{nM+2}, \dots, x_{t}), r_t \rangle \\
&\geq & \left(\sum_{t=nM + 1}^{(n+1)M} r_{t, i^*}\right) - M\cdot R(M, d).\end{eqnarray*}

Summing this over all $t \in [T]$, it follows that $\Reg(\textsc{PeriodicRestart}; \bfr) \leq R(M, d)$, as desired.


\end{proof}

As a corollary of Theorem \ref{thm:periodic_regret}, we see there exist bounded-recall algorithms with per-round regret of $O(\sqrt{(\log d)/M})$.

\begin{corollary}
For any $M > 0$, $d \geq 2$, there exists a bounded-recall algorithm $\cA$ such that for any online learning instance $\bfr \in [0, 1]^{T \times d}$, $\Reg(\cA; \bfr) \leq O(\sqrt{(\log d)/M})$.
\end{corollary}
\begin{proof}
Run \textsc{PeriodicRestart} with the \textsc{Hedge} algorithm, which has the guarantee that $\Reg(\textsc{Hedge}; \bfr) \leq \sqrt{\frac{\log d}{T}}$ for any $\bfr \in [0, 1]^{T \times d}$ (see e.g. \citet{arora2012multiplicative}). 
\end{proof}


\section{Bounded-recall mean-based algorithms have high regret}

As mentioned earlier, one of the most natural strategies for constructing a stationary bounded-recall algorithm is to run a no-regret algorithm $\cA$ of your choice on the rewards from the past $M$ rounds. In this section, we show that if $\cA$ belongs to the large class of mean-based algorithms, this does not work -- that is, we show any $M$-mean based algorithm incurs a constant amount of per-round regret.

\begin{theorem}\label{thm:main_counterexample}
Fix any $M > 0$, and let $\mathcal{A}$ be an $M$-mean-based algorithm. Then for any $T \geq 3M$, there exists an online learning instance $\bfr \in [0,1]^{2T}$ with two actions where $\Reg(\mathcal{A}; \bfr) \geq 1/18 - o(1)$. In particular, for sufficiently large $T$, $\Reg(\mathcal{A}; \bfr) = \Omega(1)$ (i.e., is at least a constant independent of $T$ and $M$).
\end{theorem}

The core idea behind the proof of Theorem \ref{thm:main_counterexample} will be the following example, where we construct an instance of length $3M$ where any $M$-mean-based algorithm incurs regret at least $\Omega(M)$.

\begin{figure}
    \centering
    \includegraphics[scale=0.25]{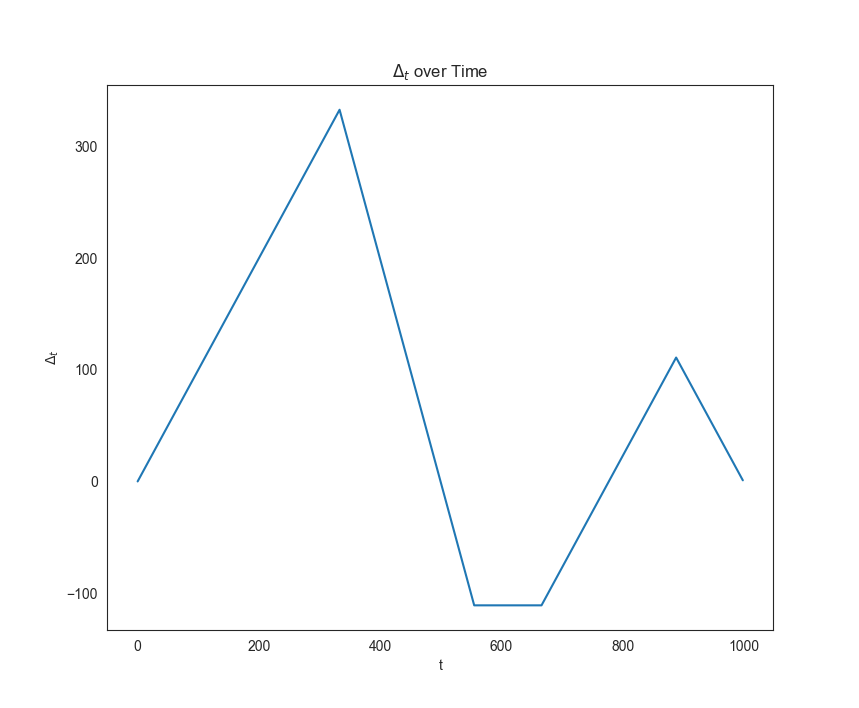}
    \caption{A plot of $\Delta_t$ over time, as used in Lemma~\ref{lem:counterexample}.}
    \label{fig:delta_t}
\end{figure}

\begin{lemma}\label{lem:counterexample}
Fix any $M > 0$, let $T = 3M$, and let $\mathcal{A}$ be an $M$-mean-based algorithm. There exists an online learning instance $\bfr \in [0, 1]^{2T}$ with two actions where $T \cdot \Reg(\mathcal{A}; \bfr) \geq M/6 - o(M)$. 
\end{lemma}
\begin{proof}
Consider the following instance $\bfr$:

\begin{itemize}
    \item For $1 \leq t \leq M$, $r_{t} = (1, 0)$.
    \item For $M < t \leq 5M/3$, $r_{t} = (0, 1)$.
    \item For $5M/3 < t \leq 2M$, $r_{t} = (1, 0)$.
    \item For $2M < t \leq 3M$, $r_{t} = (0, 0)$.
\end{itemize}

For each $t$, let $R_{t, 1} = \sum_{s = t-M}^{t-1} r_{t, 1}$ and let $R_{t, 2} = \sum_{s = t-M}^{t-1} r_{t, 2}$ (letting $r_{t} = 0$ for $t \leq 0$). Since $\mathcal{A}$ is $M$-mean-based, there exists a $\gamma$ (which is $o(1)$ w.r.t. $T$) such that if $R_{t, 1} - R_{t, 2} > \gamma T$, then $x_{t, 1} \geq 1 - \gamma$, and likewise if $R_{t, 2} - R_{t, 1} > \gamma T$, then $x_{t, 2} \geq 1 - \gamma$. Let $\Delta_{t} = R_{t, 1} - R_{t, 2}$. We then have that:

\begin{itemize}
    \item For $1 \leq t \leq M$, $\Delta_{t} = t$.
    \item For $M < t \leq 5M/3$, $\Delta_{t} = M - 2(t-M)$.
    \item For $5M/3 < t \leq 2M$, $\Delta_{t} = -M/3$.
    \item For $2M < t \leq 8M/3$, $\Delta_{t} = -M/3 + (t - 2M)$.
    \item For $8M/3 < t \leq 3M$, $\Delta_{t} = M/3 - (t - 8M/3)$. 
\end{itemize}

For a visualization, see Figure~\ref{fig:delta_t}.

Now, let $\mathcal{P} = \{t \in [T] \mid \Delta_{t} \geq \gamma T\}$, let $\mathcal{N} = \{t \in [T] \mid \Delta_{t} \leq -\gamma T\}$, and let $\mathcal{Z} = \{t \in [T] \mid t \not\in \mathcal{P} \cup \mathcal{N}\}$. As previously discussed, for $t \in \mathcal{P}$, $x_{t, 1} \geq 1 - \gamma$, and for $t \in \mathcal{N}$, $x_{t, 2} \geq 1-\gamma$. It follows that the total reward obtained by $\mathcal{A}$ over $\bfr$ is at most

\begin{equation}\label{eq:counterexample_util}
\sum_{t=1}^{T} \langle r_t, x_t \rangle \leq \left(\sum_{t \in \mathcal{P}} r_{t, 1} + \sum_{t \in \mathcal{N}} r_{t, 2}\right) + \gamma T + |\mathcal{Z}|.\end{equation}

\noindent
From our characterization above of $\Delta_t$, we know that:

$$\mathcal{P} = \left[\gamma T, \frac{3M}{2} - \frac{\gamma}{2}T\right] \bigcup \left[\frac{7M}{3} + \gamma T, 3M - \gamma T\right],$$

\noindent
and

$$\mathcal{N} = \left[\frac{3M}{2} + \frac{\gamma}{2}T, \frac{7M}{3} - \gamma T\right].$$

\noindent
Combining this with the description of the instance $\mathcal{A}$, we can see that

$$\sum_{t \in \mathcal{P}} r_{t, 1} = M - \gamma T,\; \sum_{t \in \mathcal{N}} r_{t, 2} = \frac{M}{6} - \frac{\gamma}{2} T.$$

\noindent
and that $|\mathcal{Z}| \leq 5 \gamma T$. Expression \eqref{eq:counterexample_util} for the reward of the learner then becomes

\begin{equation}\label{eq:counterexample_util2}
\sum_{t=1}^{T} \langle r_t, x_t \rangle \leq \frac{7M}{6} + \frac{9\gamma}{2}T = \frac{7M}{6} + o(T).
\end{equation}

On the other hand, the optimal action in hindsight is action $1$, and $\sum_{t=1}^{T} r_{t, 1} = \frac{4M}{3}$. It follows that

$$T \cdot \Reg(\mathcal{A}; \bfr) = \sum_{t=1}^{T} r_{t, 1} - \sum_{t=1}^{T} \langle r_t, x_t \rangle \geq \frac{M}{18} - o(M).$$
\end{proof}

\begin{remark}
It is valuable to compare the example in Lemma \ref{lem:counterexample} to the example provided by \citet{zapechelnyuk2008better} in their Theorem 4 (proving high-regret against better-reply dynamics). Like us, their example can be broken down in to a small number of phases of distinct behavior; however, they require the adversary to adapt to the learner's actions, and their example does not clearly generalize to mean-based algorithms.
\end{remark}

We now apply Lemma \ref{lem:counterexample} to prove Theorem \ref{thm:main_counterexample}. 

\begin{proof}[Proof of Theorem~\ref{thm:main_counterexample}]
Let $\bfr^{(M)} \in [0, 1]^{2 \cdot (3M)}$ be the counterexample constructed in Lemma \ref{lem:counterexample}, and let $N = \lfloor T / 3M \rfloor$. Construct $\bfr \in [0, 1]^{T}$ by concatenating $N$ copies of $\bfr^{(M)}$ and setting all other rewards to $0$ (i.e., $\bfr_{t, i} = \bfr^{(M)}_{(t\bmod 3M), i}$ for $t \leq 3MN$, $\bfr_{t, i} = 0$ for $t > 3MN$). 

Fix any $1 \leq n \leq N$, and consider the regret incurred by $\cA$ on rounds $t \in [(n-1)\cdot 3M +1, n\cdot 3M]$ (the $n$th copy of $\bfr^{(M)}$). Since $\bfr^{(M)}$ ends with $M$ zeros, $\cA$ will behave identically on these rounds as it would in the first $T$ rounds. Thus, by Lemma \ref{lem:counterexample}, $\cA$ incurs regret at least $M/6 - o(M)$ on these rounds, and at least $NM/6 - o(T)$ regret in total. The per-round regret of $\cA$ is thus at least $\Reg(\cA; \bfr) \geq \frac{1}{T}\left(\frac{NM}{6} - o(T)\right) \geq \frac{1}{18} - o(1).$
\end{proof}






\section{Stationary Bounded-Recall Algorithms}

\subsection{Averaging over restarts}

In the previous section, we showed that running a mean-based learning algorithm over the restricted history does not result in a low-regret stationary bounded-recall learning algorithm. But are there \textit{non-mean-based} algorithms which lead to low-regret stationary bounded-recall algorithms? Do low-regret stationary bounded-recall algorithms exist at all?

In this section, we will show that the answer to both of these questions is yes. We begin by constructing a stationary $M$-bounded-recall algorithm called the \textsc{AverageRestart} algorithm (Algorithm~\ref{alg:avg_restart_NR}) which incurs per-round regret of at most $O(\sqrt{(\log d)/M})$ (matching the lower bound of Theorem \ref{thm:lb}). Intuitively, \textsc{AverageRestart} randomizes over several different versions of \textsc{PeriodicRestart} -- in particular, one can view the output of \textsc{AverageRestart} as first randomly sampling a starting point $s$ uniformly over the last $M$ rounds, and outputting the action that $\cA$ would play having only seen rewards $r_{s}$ through $r_{t-1}$ (if $s \leq 0$ or $s > T$, we assume that $r_{s} = 0$). Note that this algorithm is stationary $M$-bounded-recall, as no step of this algorithm depends specifically on the round $t$.

\begin{algorithm}
\caption{\textsc{AverageRestart}}\label{alg:avg_restart_NR}
\begin{algorithmic}[1]
\REQUIRE Time horizon $T$, history window $M$, no-regret algorithm $\mathcal{A}$.
\FOR{$t = 0 \to T$} 
    \FOR{$m = 1 \to M$} 
        \STATE{$x_{t}^{(m)} := \cA(r_{t-m}, r_{t-m+1}, \dots, r_{t-1})$ (where we let $r_{t} = 0$ for $t \leq 0$)}.
    \ENDFOR
    \STATE{Play action $x_t = \frac{1}{M}\sum_{m=1}^{M} x_{t}^{(m)}$.}
\ENDFOR
\end{algorithmic}
\end{algorithm}

\begin{algorithm}
\caption{\textsc{RandomizedAverageRestart}}\label{alg:avg_restart_NR}
\begin{algorithmic}[1]
\REQUIRE Time horizon $T$, history window $M$, no-regret algorithm $\mathcal{A}$.
\FOR{$t = 0 \to T$}  
    \STATE{Sample $j \in [M]$ uniformly at random.}
    \STATE{$x_{t}^{(j)} := \cA(r_{t-j}, r_{t-j+1}, \dots, r_{t-1})$ (where we let $r_{t} = 0$ for $t \leq 0$)}.
    \STATE{Play action $x_t := x_t^{(j)}$.}
\ENDFOR
\end{algorithmic}
\end{algorithm}

\begin{theorem}
\label{thm:average_regret}
Assume the algorithm $\mathcal{A}$ has the property that $\Reg(\mathcal{A}; \bfr) \leq R(T, d)$ for any online learning instance $\bfr \in [0, 1]^{T \times d}$. Then, for any instance $\bfr \in [0, 1]^{T \times d}$

$$\Reg(\textsc{AverageRestart}; \bfr) \leq R(M, d)$$
\end{theorem}
\begin{proof}
See Appendix.
\end{proof}

By standard concentration arguments, the corresponding randomized algorithm $\textsc{RandomizedAverageRestart}$ also has low regret with high probability.

\begin{corollary}\label{cor:rand_restart}
Fix $\delta > 0$, and define $\mathcal{A}$, $R$, and $\mathbf{r}$ as in Theorem \ref{thm:average_regret}. Then, with probability at least $1-\delta$ (over the randomness due to the algorithm in all rounds), 

\begin{multline*}
\Reg(\textsc{RandomizedAverageRestart}; \bfr) \leq \\ R(M, d) + \sqrt{T\log(1/\delta)}.
\end{multline*}
\end{corollary}

As with \textsc{PeriodicRestart}, by choosing $\cA$ to be \textsc{Hedge}, we obtain a stationary $M$-bounded-recall algorithm with regret $O(\sqrt{(\log d)/M})$.

\begin{corollary}
For any $M > 0$, $d \geq 2$, there exists a stationary bounded-recall algorithm $\cA$ such that for any online learning instance $\bfr \in [0, 1]^{T \times d}$, $\Reg(\cA; \bfr) \leq O(\sqrt{(\log d)/M})$.
\end{corollary}

\subsection{Averaging restarts over the entire time horizon}

Earlier, we asked whether there were non-mean-based learning algorithms which give rise to stationary bounded-recall learning algorithms with regret $o(1)$. While it is not immediately obvious from the description of $\textsc{AverageRestart}$, this algorithm is indeed of this form. We call the corresponding (non-bounded-recall) learning algorithm \textsc{AverageRestartFullHorizon} (Algorithm~\ref{alg:avg_restart_full}). In particular, the action $x_t$ output by \textsc{AverageRestart} at time $t$ is given by

$$x_t = \textsc{AverageRestartFullHorizon}(r_{t - M}, \dots, r_{t-1}).$$

where $\textsc{AverageRestartFullHorizon}$ is initialized with time horizon $M$. 

\begin{algorithm}
\caption{\textsc{AverageRestartFullHorizon}}\label{alg:avg_restart_full}
\begin{algorithmic}[1]
\REQUIRE Time horizon $T$, no-regret algorithm $\mathcal{A}$.
\FOR{$t = 0 \to T$} 
    \FOR{$\tau = 1 \to T$} 
        \STATE{$x_{t}^{(\tau)} := \cA(r_{t-\tau}, r_{t-\tau+1}, \dots, r_{t-1})$ (where we let $r_{t} = 0$ for $t \leq 0$)}.
    \ENDFOR
    \STATE{Play action $x_t = \frac{1}{T}\sum_{\tau=1}^{T} x_{t}^{(\tau)}$.}
\ENDFOR
\end{algorithmic}
\end{algorithm}

Interestingly, if $\cA$ is a low-regret algorithm, so is \textsc{AverageRestartFullHorizon}. This follows directly from Theorem \ref{thm:average_regret}.

\begin{theorem}\label{thm:average_restart_full}
Assume the algorithm $\mathcal{A}$ has the property that $\Reg(\mathcal{A}; \bfr) \leq R(T, d)$ for any online learning instance $\bfr \in [0, 1]^{T \times d}$. Then, for any instance $\bfr \in [0, 1]^{T \times d}$

$$\Reg(\textsc{AverageRestartFullHorizon}; \bfr) \leq R(T, d).$$
\end{theorem}
\begin{proof}
Set $M = T$ in the proof of Theorem \ref{thm:average_regret}. (In particular, the proof of Theorem \ref{thm:average_regret} applies for any choice of $M$ and $T$, not only $T \gg M$). 
\end{proof}

\subsection{The necessity of asymmetry}

One basic (and somewhat surprising) property of many mean-based algorithms is that they treat all rounds in the past symmetrically: permuting the order of the past rounds does not affect the action chosen by Hedge or FTRL. Mathematically, this is equivalent to saying that the function $\cA(r_1, r_2, \dots, r_{t-1})$ is a symmetric function in its arguments.

Likewise, we can say that a bounded-recall algorithm is \emph{symmetric} if its action $x_t$ at time $t$ is a symmetric function of the rewards $r_{t-M}, \dots, r_{t-1}$ (i.e., the functions $f_t$ in Definition \ref{def:bounded-recall} are symmetric). Given the prevalence of symmetric no-regret learning algorithms, it is natural to ask whether there exist any symmetric no-regret bounded-recall learning algorithms. Notably, both the periodic restart algorithm and average restart algorithm are not symmetric (they both put significantly more weight on recent rewards).

In this section, we prove that there does not exist a bounded-recall learning algorithm that is all three of stationary, symmetric, and no-regret. We will do this by showing that any such learning algorithm must act ``similarly'' to a mean-based algorithm. This will allow us to use Lemma \ref{lem:counterexample} to show that any symmetric bounded-recall algorithm must have high-regret.

The key lemma we employ is the following.

\begin{lemma}\label{lem:symm-is-mb}
Let $\cA$ be a stationary, symmetric $M$-bounded-recall no-regret learning algorithm, and choose a sufficiently large $T \geq 10M/\eps^2$ such that $\Reg(\cA) \leq \eps/10$. For any $0 \leq n \leq M$, let $p(n)$ be the probability that $\cA$ plays arm $1$ if, of the $M$ previous rewards $r_{t-M}, \dots r_{t-1}$, exactly $n$ of them are equal to $(1, 0)$ and exactly $M-n$ of them are equal to $(0, 1)$. 

Then, if $n \leq (1-\eps)(M/2)$, $p(n) \leq \eps$, and if $n \geq (1+\eps)(M/2)$, $p(n) \geq 1-\eps$.
\end{lemma}
\begin{proof}[Proof sketch]
The main idea is to construct a sequence of rewards $\bfr$ where any segment of $M$ consecutive rewards in $\bfr$ has the property that exactly $n$ of them equal $(1, 0)$, and $M-n$ of them equal $(0, 1)$. It is possible to construct such a sequence by taking a segment of $M$ rewards with this property and repeatedly cycling it.

For such a sequence of rewards, $\cA$ is forced to select arm 1 with probability exactly $p(n)$ for each round $t > M$. Examining the regret of $\cA$ on this sequence leads to the lemma statement; we defer details to the Appendix.
\end{proof}

Lemma \ref{lem:symm-is-mb} can be thought of as satisfying the following ``weak'' form of the mean-based condition: whenever most of the previous rewards are $(1, 0)$, $\cA$ must put the majority of its weight on arm $1$, and whenever most of the previous rewards are $(0, 1)$, $\cA$ must put the majority of its weight on arm $2$. But since the counterexample in Lemma \ref{lem:counterexample} primarily uses rewards of the form $(1, 0)$ and $(0, 1)$, this weak condition is sufficient to prove an analogue of Theorem \ref{thm:main_counterexample}.

\begin{theorem}
Let $\cA$ be an $M$-mean-based symmetric, stationary, learning algorithm. Then, for any sufficiently large $T$ there exists an online learning instance $\bfr \in [0,1]^{2T}$ with two actions where $\Reg(\mathcal{A}; \bfr) = \Omega(1)$.
\end{theorem}
\begin{proof}
We first claim that if we take $T$ large enough so that Lemma \ref{lem:symm-is-mb} is satisfied, then $\cA$ will incur regret at least $(1-O(\eps)) \cdot (M/18) - o(M)$ on a segment of $3M$ rounds constructed as in Lemma \ref{lem:counterexample}. To see this, note that the analysis of Lemma \ref{lem:counterexample} holds essentially as written, with the exception that we can replace the set $\cP$ with $\cP_{\eps} = \{t \in [T] | \Delta_t \geq \gamma T + \eps(M/2)\}$, with the guarantee that $x_{t, 1} \geq 1 - \gamma - \eps$ for $t \in \cP_{\eps}$. Likewise, we can replace $\cN$ with the analogous set $\cN_{\eps}$ with the analogous guarantee. Both these changes change the LHS of \eqref{eq:counterexample_util2} by at most $O(\eps)$, and we therefore arrive at the above regret bound.

(There is one subtle issue with the above argument, which is that for the first $M$ rounds of this segment, some of the rewards in the history window are $(0, 0)$. But for these first $M$ rounds, the mean-based algorithm plays optimally, so misplaying here cannot decrease our regret.)

Now, by repeating the same concatenation argument as in the proof of Theorem \ref{thm:main_counterexample}, we can show that $\cA$ must incur a per-round-regret of at least $1/18 - O(\eps) - o(1)$ on $\lfloor T/3M \rfloor$ concatenated instances of the above segment. 
\end{proof}

\section{Simulations}

\begin{figure*}
    \centering
    \begin{subfigure}{.5\textwidth}
    \centering
    \includegraphics[scale=0.37]{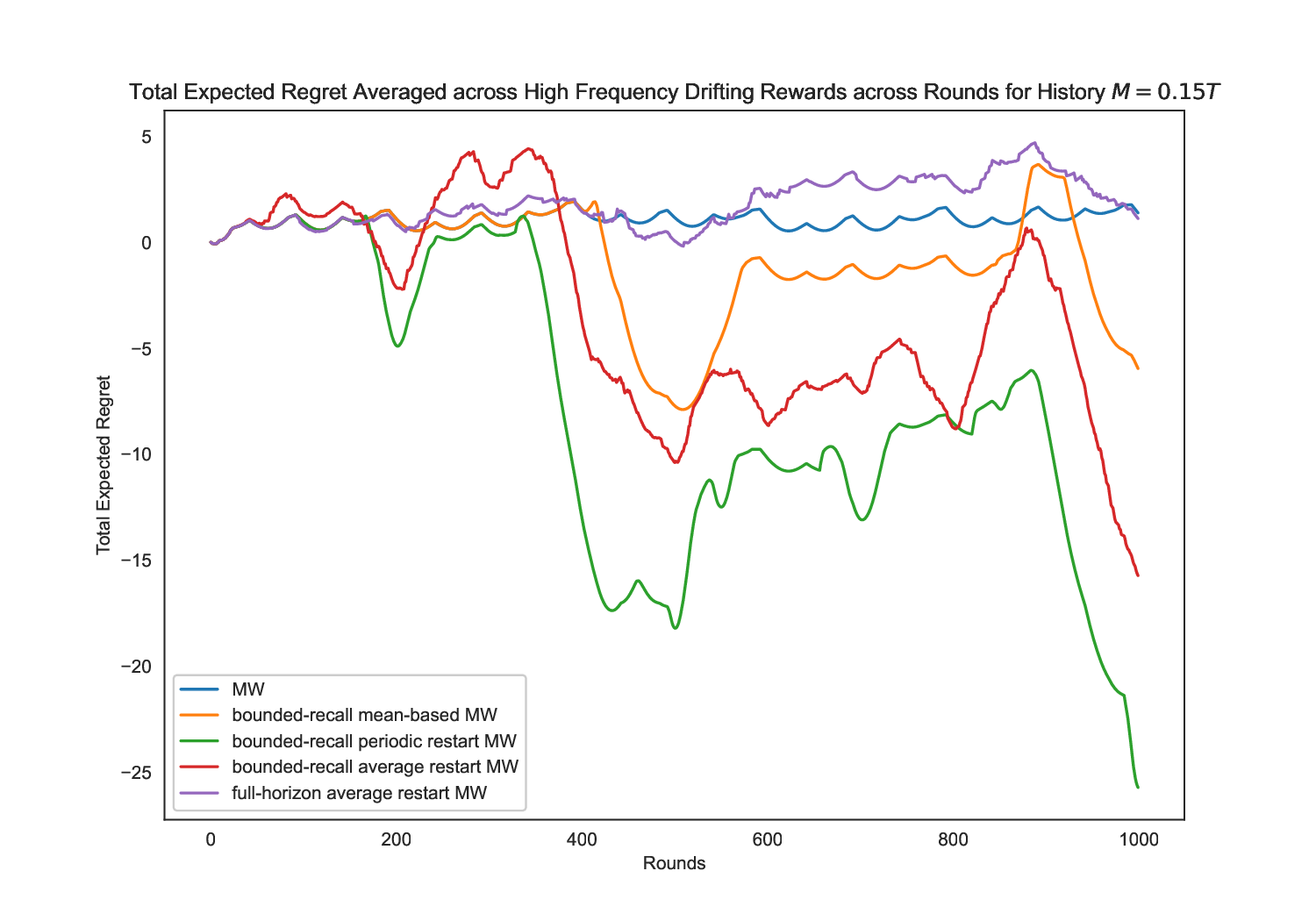}
    \caption{Left}
    \end{subfigure}%
    \begin{subfigure}{.5\textwidth}
    \centering
    \includegraphics[scale=0.37]{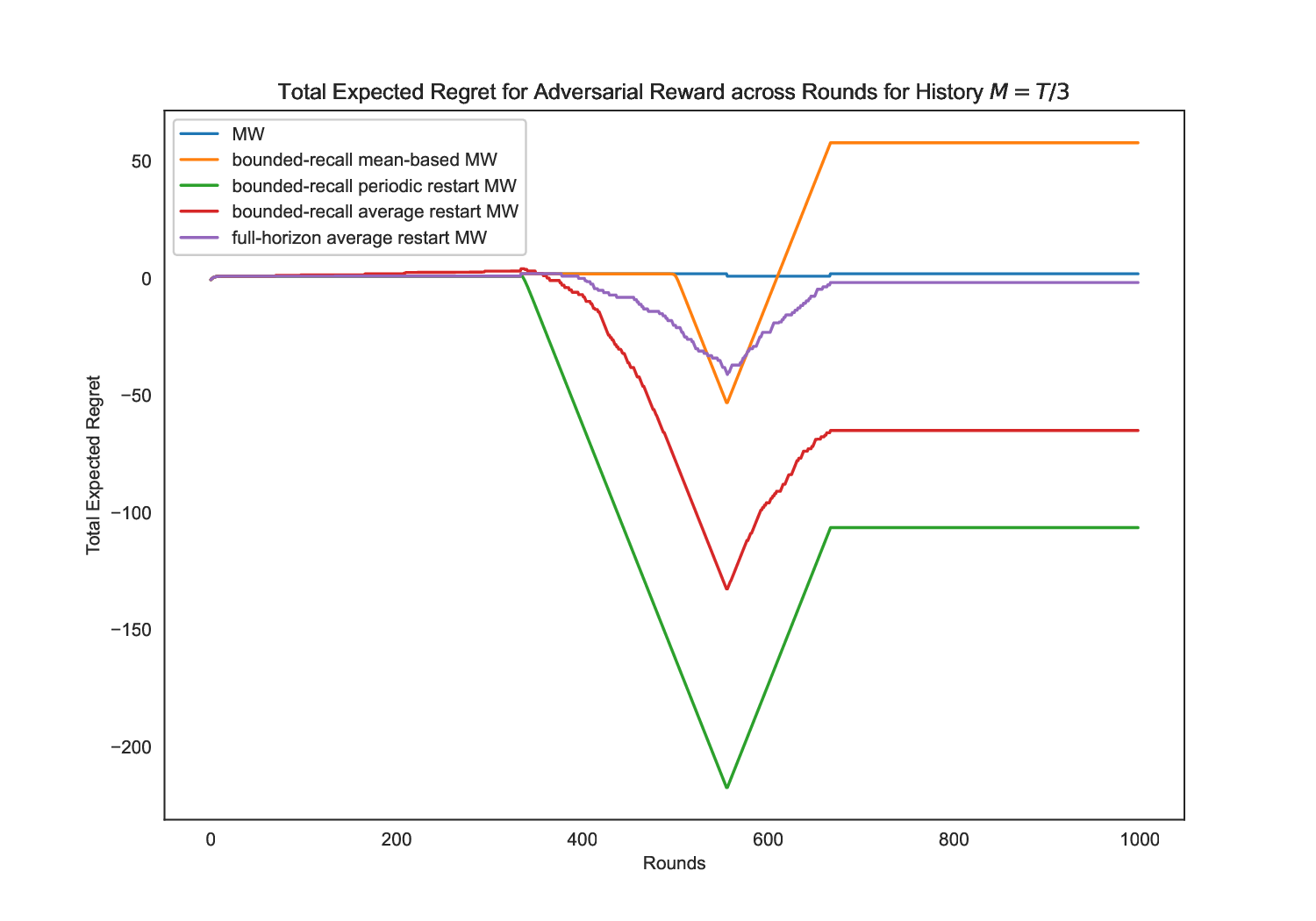}
    \caption{Right}
    \end{subfigure}
    \caption{(Left) We plot the total regret of the algorithms over time over a uniform average of high-frequency drifting scenarios where the periods of the mean reward of arm $1$ are $T/20, T/10, T/5,$ and $T/2$ and arm $2$ flips an unbiased coin for reward $\{\pm 1\}$ -- the bounded-recall algorithms significantly outperform the classic no-regret algorithms. (Right) We plot the total regret of the algorithms over time for one block of the adversarial rewards case (see the construction in Lemma~\ref{lem:counterexample}) -- observe that the mean-based bounded-recall learner attains regret on order $M/6$ (here, $M = T/3$), while our no-regret bounded-recall learners all outperform Multiplicative Weights.}
    \label{fig:total_regret_vs_time}
\end{figure*}


We conduct some experiments to assess the performance of the bounded-recall learning algorithms we introduce in some simple settings. In particular, we consider Algorithms~\ref{alg:restart_NR} (\textsc{PeriodicRestart}), ~\ref{alg:avg_restart_NR} (\textsc{AverageRestart}), ~\ref{alg:avg_restart_full} (\textsc{AverageRestartFullHorizon}), and an $M$-bounded-recall mean-based algorithm, all based off of the Multiplicative Weights Update algorithm with fixed learning rate $\eta = 1/2$. We also simulate the classic full-horizon Multiplicative Weights algorithm with the same parameters. 

We assess performance in two synthetic environments. In the first environment, we simulate periodic drifting rewards, where the expected reward of arm 1 at time $t$ has mean $|\sin(\pi/6 + t \cdot \pi/\phi)|$, for a period $\phi$ uniformly chosen from the set $\{T/20, T/10, T/5, T/2\}$. All algorithms we simulate in the first environment have $M = 0.15T$. In the second environment, we simulate the adversarial example from the proof of Lemma \ref{lem:counterexample} (with $M = T/3$). In Figure \ref{fig:total_regret_vs_time}, we plot the cumulative regret over time for both environments, averaged over ten independent runs with $T = 1000$.

As we might expect, the bounded-recall algorithms outperform the full-horizon algorithms in the first environment, by virtue of being able to quickly respond to the recent past. In the second environment, we see that, confirming the claims of Theorem \ref{thm:main_counterexample}, the choice of bounded-recall algorithm can have a large impact on regret: although \textsc{PeriodicRestart} and \textsc{AverageRestart} end with significant negative regret, the naive mean-based bounded-recall algorithm ends up with significant positive regret (proportional to $T$). Interestingly, the bounded-recall no-regret algorithms also outperform the full-horizon no-regret algorithms here (which both end with approximately zero regret).

\section{Conclusion and Future Work}

In this paper, we initiated the study of bounded-recall online learning and produced novel bounded-recall no-regret algorithms, as well as provided linear regret lower bounds against a natural class of bounded-recall learners. 

There are many avenues for future work:
\begin{itemize}
    \item It would be interesting to consider bounded-recall extensions to other notions of regret (e.g., swap regret \citep{blum2007external});
    \item Our stationary bounded-recall algorithms are more computationally demanding than other no-regret algorithms -- understanding the computational limitations of these methods would be useful as well;
    \item Obtaining a theoretical characterization of performance improvements for bounded-recall methods relative to classic online learners in general non-stationary settings would be quite interesting as well.
\end{itemize}

\section*{Impact Statement}

This paper presents work whose goal is to advance the field of 
Machine Learning. There are many potential societal consequences 
of our work, none which we feel must be specifically highlighted here.

\bibliography{main}
\bibliographystyle{icml2024}

\newpage
\appendix
\onecolumn

\section{Related Work}
\label{sec:related_work}

\subsection{Adaptive Multiplicative Weights}
\looseness=-1
A lot of existing work focuses on understanding the behavior of Multiplicative Weights under various non-adversarial assumptions and properties of the loss sequence \citep{cesa2007improved, hazan2010extracting}, as well as on coming up with adaptive algorithms which perform better than Multiplicative Weights in both worst-case and average-case settings \citep{erven2011adaptive, de2014follow, koolen2016combining, mourtada2019optimality}. There is a particular emphasis on trading off between favorable performance for Follow-the-Leader and Multiplicative Weights in various average case settings. Our work connects to this literature by introducing bounded-recall algorithms which empirically outperform both MW and FTL for a class of drifting and periodic rewards.

\subsection{Private Learning and Discarding Data}

Bounded-recall online learning algorithms can be viewed as private with respect to data sufficiently far in the past -- such data is not taken into account in the prediction. This approach to achieving privacy is similar in principle to \textit{federated learning} \citep{kairouz2021advances}, which ensures that by decentralizing the data store, many parties participating in the model training will simply never come into contact with certain raw data points, thus mitigating privacy risks. Similarly, bounded-recall approaches also provide an alternate angle on privacy-preserving ML systems which must store increasing amounts of streaming data (and thereby must adaptively set their privacy costs to avoid running out of privacy budget, see \citet{lecuyer2019privacy}). Additionally, the bounded-recall approach to privacy yields algorithms for which data deletion \citep{ginart2019making} is efficient -- thereby ensuring ``the right to be forgotten.'' 






\subsection{Shifting Data Distributions}

Bounded-recall algorithms are a natural approach to online learning over non-stationary time series, which is a common problem in practical industry settings \citep{huyen2022data} and which has been studied for many years \citep{sugiyama2012machine, wiles2021fine, wu2021learning, rabanser2019failing}. In particular, one can view bounded-recall online learners as adapting to non-stationary structure in the data -- thus, one may not need to go through the whole process of detecting a distribution shift and then deciding to re-train -- ideally the learning algorithm is adaptive and automatically takes such eventualities into account. Our proposed bounded-recall methods are one step in this direction.







\section{Omitted Proofs}
\label{sec:omitted_proofs}

\subsection{Proof of Theorem~\ref{thm:lb}}
\begin{proof}[Proof of Theorem~\ref{thm:lb}]
Our hard example is simple: we make use of standard hard examples used in regret lower bounds for the classical online learning setting (see e.g. \citet{roughgarden2016online}), and simply append to such an example of length $M$ a block of $M$ $0$ rewards for all actions (effectively resetting the internal state of any bounded-recall algorithm, and resulting in $0$ regret during that block). This trick allows us to only consider the regret on blocks of size $M$, and since the blocks are repeated, each block has the same best action in hindsight. Then taking the full block of length $2M$ together, we get an average regret lower bound for each block of size $\frac{1}{2}\Omega(\sqrt{M \log d})$. Adding up the regret lower bounds, we get a total regret lower bound for any bounded-recall online learning algorithm with past window of size $M$ to be $\Omega\left(\frac{T}{2M}\cdot \sqrt{M \log d}\right) = \Omega\left(\sqrt{\frac{T^2 \log d}{M}}\right)$, or $\Omega\left(\sqrt{\frac{\log d}{M}}\right)$ on average, as desired.
\end{proof}

\subsection{Proof of Theorem~\ref{thm:average_regret}}
\begin{proof}[Proof of Theorem~\ref{thm:average_regret}]
We will proceed by first proving the statement for the deterministic algorithm, the proof for the randomized algorithm follows directly. Intuitively, we will decompose the output of \textsc{AverageRestart} as a uniform combination of $M$ copies of \textsc{PeriodicRestart} (one for each offset modulo $M$ of reset location); since \textsc{PeriodicRestart} has per-round regret $R(M, d)$, so will \textsc{AverageRestart}. 

Let $i^* = \arg\max_{i \in [d]} \sum_{t} r_{t, i}$. For any $t \in [-(M-1), T-1]$ and $m \in [M]$, let $y_{t}^{(m)} = x_{t+m}^{(m)} = \cA(r_{t}, r_{t+1}, \dots, r_{t+m-1})$. Now, note that

\begin{eqnarray*}
\sum_{t=1}^{T} \langle x_{t}, r_{t} \rangle &=& \sum_{t=1}^{T}\frac{1}{M}\sum_{m=1}^{M} \langle x_{t}^{(m)}, r_{t} \rangle\\
&=& \sum_{t=1}^{T}\frac{1}{M}\sum_{m=1}^{M} \langle y_{t-m}^{(m)}, r_{t} \rangle \\
&=& \sum_{t'=-M+1}^{T-1}\frac{1}{M}\sum_{m=1}^{M} \langle y_{t'}^{(m)}, r_{t' + m} \rangle \\
&=& \sum_{t'=-M+1}^{T-1}\frac{1}{M}\sum_{m=1}^{M} \langle \cA(r_{t'}, r_{t'+1}, \dots, r_{t' + m-1}), r_{t' + m} \rangle \\
&\geq & \sum_{t'=-M+1}^{T-1}\frac{1}{M}\left(\sum_{m=1}^{M} r_{t'+m, i^*} - M\cdot R(M, d)\right) \\
&=& \left(\sum_{t=1}^{T} r_{t, i^*}\right) - (T + M)R(M, d).
\end{eqnarray*}

Here we have twice used the fact that $r_{s} = 0$ for $s \leq 0$ and $s > T$; once for rewriting the sum in $t$ in terms of $t'$ (the terms that do not appear in the original sum have $t' + m \not \in [1, T]$, so the inner product evaluates to $0$), and once again when we rewrite the sum in $t'$ in terms of $T$ (each $r_{t, i}$ for $1 \leq t \leq T$ appears exactly $M$ times; other $r_{t, i}$ for $t \not\in [1, T]$ appear a variable number of times, but they all equal $0$). Since $M \leq T$, it follows that $\Reg(\textsc{AverageRestart}; \bfr) \leq R(M,d)$. 

\end{proof}

\subsection{Proof of Corollary~\ref{cor:rand_restart}}

\begin{proof}[Proof of Corollary~\ref{cor:rand_restart}]

To extend this proof to the randomized variant of the algorithm, note that the expected reward of the randomized algorithm in round $t$ is equal to the reward of the deterministic algorithm via linearity of expectation:
\begin{eqnarray*}
    \mathop{{}\mathbb{E}}_{j \sim \textnormal{Unif}([M])}\left[\sum_{t=1}^{T}\langle x_{t}^{(j)}, r_{t} \rangle\right] &=& \sum_{t=1}^{T}\mathop{{}\mathbb{E}}_{j \sim \textnormal{Unif}([M])}\left[\langle x_{t}^{(j)}, r_{t} \rangle\right] = \sum_{t=1}^{T}\frac{1}{M}\sum_{m=1}^{M} \langle x_{t}^{(m)}, r_{t} \rangle.
\end{eqnarray*}

Now, let $X_t = \langle x_{t}^{(j)}, r_{t} \rangle$ be the r.v. corresponding to the randomized algorithm's reward in round $t$, and $\overline{X}_t = \mathbb{E}[X_t] = \frac{1}{M}\sum_{m=1}^{M} \langle x_{t}^{(m)}, r_{t} \rangle$ the reward of the deterministic algorithm in round $t$. Since $x_t \in \Delta_d$ and $r_t \in [0, 1]^d$, $X_t$ lies in $[0, 1]$, so by Hoeffding's inequality

$$\Pr\left[\sum X_t - \sum \overline{X}_t \geq - \sqrt{T \log(1/\delta)}\right] \geq 1 - \delta,$$

as desired.
\end{proof}

\subsection{Proof of Lemma~\ref{lem:symm-is-mb}}

\begin{proof}[Proof of Lemma~\ref{lem:symm-is-mb}]
We will assume that $n \leq (1-\eps)(M/2)$ (the other case can be handled symmetrically). Fix any sequence $\bfr_{M}$ of $M$ rewards, $n$ of which are $(1, 0)$ and $M - n$ of which are $(0, 1)$. Form a sequence $\bfr$ of $T$ rewards by repeating $\bfr_{M}$ multiple times. 

Note that $\bfr$ has the property that any segment of $M$ consecutive rewards contains exactly $n$ $(1, 0)$s and $M-n$ $(0, 1)$s. So in each round after round $M$, the algorithm $\cA$ will play arm $1$ with probability $p(n)$. By doing this, they receive reward at most:

$$\mathrm{Reward}(\cA) \leq M + \left(p(n)\frac{n}{M} + (1-p(n))\cdot\left(1 - \frac{n}{M}\right)\right)(T - M),$$

\noindent
(since they receive at most reward 1 each round for the first $M$ rounds). Since $n < M/2$, the best fixed arm in hindsight is arm 2, so the optimal static adversary receives reward at least

$$\mathrm{Opt}(\cA) \geq \left(1 - \frac{n}{M}\right)(T - M).$$

It follows that the regret of $\cA$ on this sequence is at least

\begin{eqnarray*}
\Reg(\cA; \bfr) &=& \frac{1}{T}(\mathrm{Opt}(\cA) - \mathrm{Reward}(\cA)) \\
&\geq& \frac{1}{T}\left[(T-M)\left(1-\frac{2n}{M}\right)p(n) - M\right]\\
&\geq& (1 - 0.1\eps^2)(\eps)p(n) - 0.1\eps \\
&\geq& 0.5\eps p(n) - 0.1 \eps^2
\end{eqnarray*}

Since $\Reg(\cA) \leq 0.1\eps^2$, this implies that $p(n) \leq (0.2 \eps^2)/(0.5 \eps) \leq \eps$, as desired.
\end{proof}


\end{document}